\pdfoutput=1

\documentclass[11pt]{article}

\usepackage{acl}

\usepackage{times}
\usepackage{latexsym}

\usepackage[T1]{fontenc}

\usepackage[utf8]{inputenc}

\usepackage{microtype}
\usepackage[normalem]{ulem}
\usepackage{times}
\usepackage{latexsym}
\usepackage{booktabs}
\usepackage[colorinlistoftodos]{todonotes}
\usepackage{enumitem}
\usepackage{xspace}
\usepackage{color, colortbl}
\usepackage{amssymb}
\usepackage{pifont}
\usepackage{tablefootnote}

\newcommand{\cmark}{\ding{51}}%
\newcommand{\xmark}{\ding{55}}%

\newcommand{\Thead}[1]{\textsc{\textbf{#1}}}

\newcommand{\model}{{\fontfamily{lmtt}\selectfont ePiC}\xspace}
\newcommand{\secref}[1]{\S\ref{#1}}
\definecolor{LightCyan}{RGB}{172,204,186}

\title{\model: Employing Proverbs in Context as a Benchmark for Abstract Language Understanding}

\author{Sayan Ghosh \unskip\enspace{\rm}\enspace Shashank Srivastava\\
  UNC Chapel Hill\\ 
  \texttt{\{sayghosh,ssrivastava\}@cs.unc.edu}
}

\begin{document}
\maketitle
\begin{abstract}
While large language models have shown exciting progress on several NLP benchmarks, evaluating their ability for complex analogical reasoning remains under-explored. Here, we introduce a high-quality crowdsourced dataset of narratives for employing proverbs in context as a benchmark for abstract language understanding.  
The dataset provides fine-grained annotation of aligned spans between proverbs and narratives, and contains minimal lexical overlaps between narratives and proverbs, ensuring that models need to go beyond surface-level reasoning to succeed. We explore three tasks: (1) proverb recommendation and alignment prediction, (2) narrative generation for a given proverb and topic, and (3) identifying narratives with similar motifs. Our experiments show that neural language models struggle on these tasks compared to humans, and these tasks pose multiple learning challenges.
\end{abstract}

\section{Introduction}
Large language models (LLMs)~\cite{bert, roberta, t5, bart, reimers2019sentence, distilbert, albert} have led to a paradigm shift in NLP, and have shown exciting progress on benchmarks such as GLUE~\cite{wang2018glue} and SuperGLUE~\cite{superglue}. In particular, these include tasks such as reading comprehension, natural language inference, and coreference resolution. Many of these tasks rely on semantic and syntactic reasoning, which has been mastered by these LLMs. For example, apart from improving on distributional semantics through contextualized embeddings~\cite{ethayarajh-2019-contextual}, recent work has shown evidence that these models implicitly learn emergent concepts such as subject-verb agreement~\cite{jawahar-etal-2019-bert}, semantic roles~\cite{tenney-etal-2019-bert} and dependency structures~\cite{hewitt-manning-2019-structural}. 

\begin{figure}[t]
    \centering
    \includegraphics[scale=0.42]{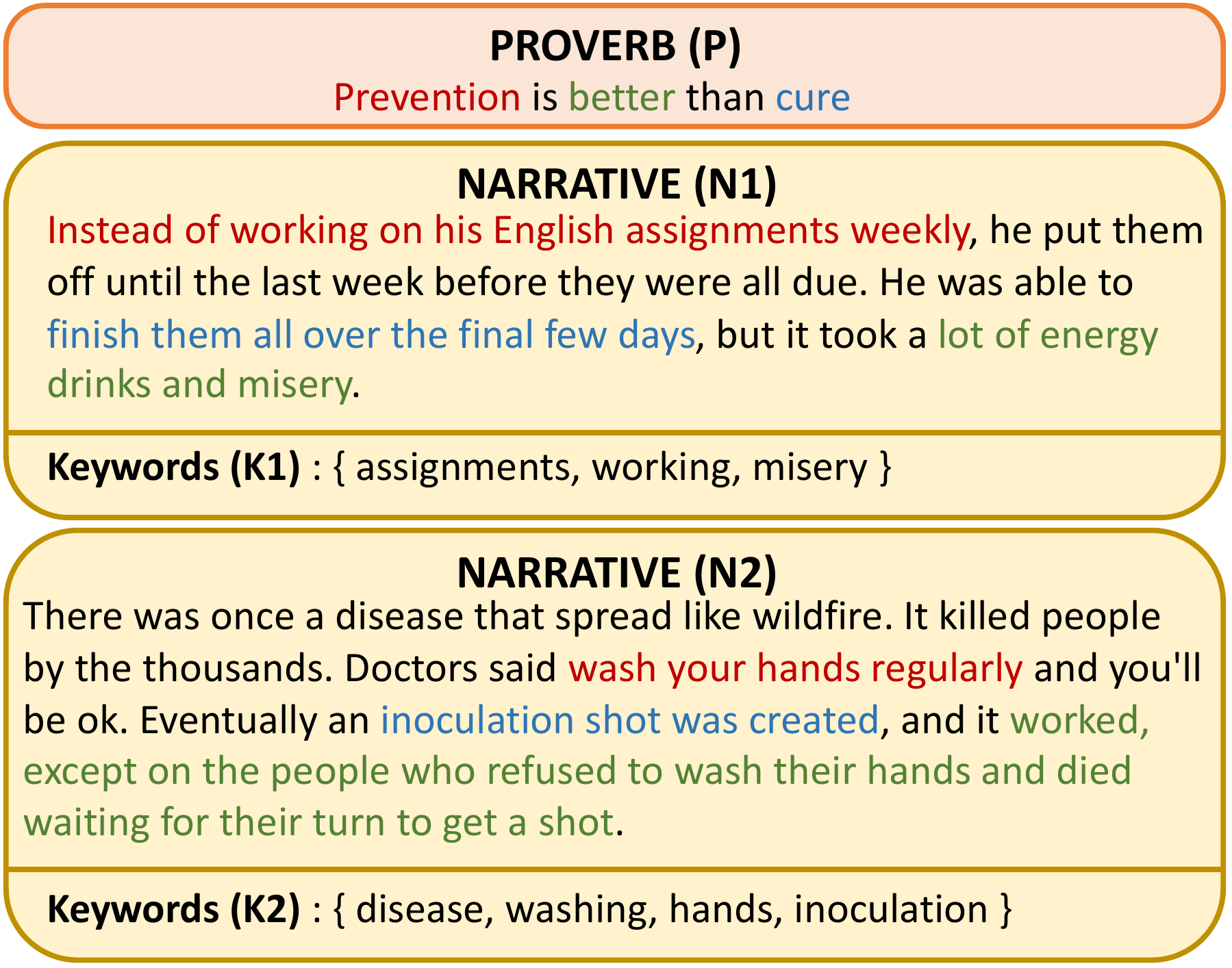}
    \caption{We introduce \model, a crowdsourced dataset of narratives for employing proverbs in context. Our dataset contains narratives (N1 and N2) paired against proverbs (P) along with a fine-grained annotation of \textit{aligned spans} between the narratives and proverbs. Aligned spans are shown with matching colors and indicate correspondences in roles between proverbs and narratives. 
    We explore three tasks: (1) proverb recommendation and alignment prediction (predict P given N1), (2) narrative generation for a given proverb and topic (generate N1 given P and K1), and (3) identifying narratives with similar motifs (e.g. identify N2 in a set of narratives given N1).}
    \label{fig:main_figure}
\end{figure}
However, humans show an ability for deeper linguistic reasoning. We can identify people's intentions and goals~\cite{douglas2006you}, perform relational reasoning~\cite{alexander2016measuring}, and find analogies in situations with little surface overlap~\cite{holyoak2013analogy}. In particular, making verbal analogies in the form of proverbs is noted as an indicator of 
literary ability~\cite{penfield1988proverbs,nippold2001proverb}. Proverbs are also repositories of information on culture, societal norms, values, and folk wisdom \cite{raymond1956tensions,white1987proverbs}. 
In this work, we investigate proverbs in narrative contexts as a testbed for evaluating abstract reasoning and analogical abilities of LLMs. 

We introduce \model $\;$(employing Proverbs in Context), a high-quality crowdsourced dataset of narratives paired with proverbs. The dataset provides fine-grained annotation of aligned spans between proverbs and narratives, and is designed to minimize lexical overlap between narratives and proverbs. Figure~\ref{fig:main_figure} shows two examples of narratives for a proverb from our dataset, along with corresponding alignment annotations. We diverge from related extant resources \cite{quote2020emnlp, ltq1, ltq2} on using proverbs in terms of quality of narratives, direct supervision, and having fine-grained alignment annotations.\footnote{Existing datasets are automatically created by scraping web-text, and supervision is heuristic (based on co-occurrences of proverbs and contexts)} We explore three tasks: (1) proverb and alignment prediction (\S~\ref{sec:proverb_retrieval}), (2) narrative generation for a given proverb and a set of keywords specifying a topic (\S~\ref{sec:narrative_generation}), and (3) discovering narratives with similar motifs (\S~\ref{sec:discovering_similar_narratives}). By benchmarking several LLMs, we find that existing models struggle with these tasks, suggesting much scope of improvement in abstract reasoning. In particular, humans show much higher performance in many cases. 

In \secref{sec:dataset_creation}, we describe the crowdsourced creation of 
the \model dataset. 
In \secref{sec:data_analysis}, we analyze lexical overlap, 
biases, and narrative quality in \model. 
\secref{sec:experiments} describes the three tasks and details of experimental evaluation of LLMs for each task. We conclude with a discussion, 
and a statement of ethics and broader impact relevant to our work. Our contributions are: 
\begin{itemize}[noitemsep, topsep=0pt, leftmargin=*]
    \item We introduce \model, a high-quality dataset for employing proverbs in context. It contains multiple narratives for English proverbs and fine-grained annotation of aligned spans between them.
    \item We design three challenging tasks 
    that require models to go beyond surface-level reasoning and provoke research towards making more socially grounded NLP systems.
    \item We benchmark the performance of several state-of-the-art large language models in our proposed tasks using our dataset. 
\end{itemize}

Our dataset and code are publicly available at: \url{https://epic-benchmark.github.io}

\section{Related Work}
Prior works in figurative language understanding have explored a diverse set of topics, such as simile detection and generation~\cite{niculae2014brighter, mpouli2017annotating, zeng2020neural,chakrabarty2020generating}, metaphor detection and generation~\cite{dagan2005pascal,gao2018neural,stowe2019linguistic, stowe2021metaphor,chakrabarty2021mermaid}, pun identification~\cite{poliak2018collecting,miller2016towards}, and quote/proverb recommendation~\cite{ltq1,ltq2,quote2020emnlp}.
Recent work~\cite{chakrabarty2021s} has also focused on interpreting idioms and similes in narratives. 
\citet{liu-etal-2019-neural} has explored recommending Chinese idioms through context-based recommendation and \citet{zheng-etal-2019-chid} formulated idiom recommendation as cloze-style reading comprehension task. 
Learning to quote has been explored based on fiction \cite{ltq1,ltq2} and
noisy social media conversations from Twitter, Reddit or Weibo \cite{ltq3, quote2020emnlp}. In the most related prior work, authors explore a quote retrieval task borrowing inspiration from context based recommendation systems \cite{huang2012recommending, he2010context}.
\citet{quote2020emnlp} formulated learning to quote as a generation task by using topic modeling \cite{ntm1, ntm2} 
in a sequence-to-sequence network.
While previous work has considered idioms, proverbs and common phrases as quotes, we specifically work with proverbs. Compared to earlier datasets, our dataset is manually created and labeled. Further, \model includes fine-grained annotations aligning parts of proverb to parts of the narrative, which has significant possibilities for model training, evaluation and interpretability.

\section{Dataset Creation}
\label{sec:dataset_creation}
In this section, we describe the steps involved in creating the dataset in detail.

\noindent \textbf{Proverb collection:} We obtained a candidate set of English proverbs by scraping websites of `The Phrase Finder'\footnote{\url{https://www.phrases.org.uk/}} and WikiQuotes\footnote{\url{https://en.wikiquotes.org/wiki/English_proverbs}}. Next, this set was manually pruned to remove lexical variations of the same proverb. 
This manual curation led to a set of 250 proverbs, which we consider in the current version of our dataset. 

\noindent \textbf{Narrative collection:} In the second step, we use Amazon Mechanical Turk to collect a diverse set of narratives corresponding to each proverb. We collect 10 narratives contributed by distinct turkers for each proverb, leading to a total of 2500 proverb-narrative pairs. We also ensure that no turker contributes a large number of narratives to alleviate annotator bias~\cite{gevanew} (where models can overfit to annotator characteristics) while encouraging diversity in writing style and content. The turkers were asked to write short realistic stories, preferably within 100 words. Additionally, to avoid surface-form biases, turkers were encouraged to minimize lexical overlap and to not mention the proverb or parts of it in the narrative. This was done so that doing well on the tasks requires a detailed understanding of the narratives rather than relying on surface-level cues. Turkers were paid 50 cents for each narrative for this task.

\noindent \textbf{Span alignment annotation:} Next, we solicit fine-grained annotations between the narratives and the proverb in form of aligned spans. For this, we present proverb-narrative pairs to turkers asking them to find contiguous spans in the narrative which align well with contiguous spans in the proverb. Turkers could submit up to 5 pairs of aligned spans per proverb-narrative pair. These aligned spans highlight the grounding of a proverb in the narrative (see Figure~\ref{fig:main_figure}). These annotations can help to verify the reasoning capabilities of various neural models by checking if these models are able to identify these correspondences, and add interpretability to our tasks. Turkers were paid 25 cents for each proverb-narrative pair annotation for this task.

\noindent \textbf{Statistics:} Table \ref{tab:dataset_collection_stats} shows the statistics of narrative collection for the proverbs. The narrative writing task was perceived as challenging yet interesting by most turkers due to (a) not having outlines about topics for the narrative beforehand (b) requirement of low lexical overlap with the proverb. Thus, the narrative writing task had a learning curve and some of the narratives submitted initially were not included in the dataset. 
\addtolength{\tabcolsep}{-3pt} 
\begin{table}[!hbt]
\small
    \centering
    \begin{tabular}{lr}
    \toprule
         \# submitted narratives & 2561\\
        \# approved narratives & 2500 \\
        \# workers participated & 166 \\
        Avg. \# approved narratives per turker & 15.06\\
        Max \# approved narratives by one turker & 168 \\
    \bottomrule
    \end{tabular}
    \caption{Statistics of AMT task for narrative collection.}
    \label{tab:dataset_collection_stats}
\end{table}
\addtolength{\tabcolsep}{3pt} 
\begin{table}[!hbt]
\small
    \centering
    \begin{tabular}{lr}
    \toprule
        Vocabulary size & 16170\\
        Avg. no. of tokens per narrative & 64.27 \\
        Avg. no. of sentences per narrative & 4.26 \\
        Avg. no. of aligned spans & 2.18 \\ 
        Avg. no. words per proverb span & 2.71 \\
        Avg. no. words per narrative span & 11.57 \\
        No. of unique bigrams & 80978 \\
        No. of unique trigrams & 133772 \\
        \bottomrule
    \end{tabular}
    \caption{Dataset statistics for \model.}
    \label{tab:dataset_stats}
\end{table}
\section{Dataset Analysis}
\label{sec:data_analysis}
Table \ref{tab:dataset_stats} shows some statistics of the dataset collected through the process described in \secref{sec:dataset_creation}. 
In this section, we analyze the characteristics and biases of the \model dataset in detail. 

\begin{table}[!hbt]
\small
    \centering
    \begin{tabular}{l|c|c}
    \toprule
    \Thead{N-gram} & \Thead{Jaccard Sim.} & \Thead{Common N-grams}\\
    \midrule
         unigram & 0.0258 (0.0211) & 1.27 (1.06) \\
         bigram & 0.0010 (0.0004) & 0.07 (0.03) \\
         trigram & 0.0003 (0.0000) & 0.02 (0.00)\\
         \bottomrule
    \end{tabular}
    
    \caption{Avg. Jaccard similarity and number of common n-grams between proverbs and narratives. Numbers in parenthesis denote the corresponding statistics upon random assignment of proverbs to narratives.
    }
    \label{tab:jaccard}
\end{table}

\subsection{Lexical overlap analysis}
\label{sec:lexical}
\textbf{Using n-grams}:
We evaluate the extent of lexical overlap between proverbs and narratives by computing common n-grams between them. Table~\ref{tab:jaccard} reports the average Jaccard similarity score between n-gram sets of proverbs and narratives, and the average number of common n-grams. 
On average, there are 1.27 unigrams common between narratives and proverbs (including stopwords). In comparison, randomly permuting assignments of proverbs for narratives yields an average unigram Jaccard similarity of 0.0211 and 1.06 common unigrams. Thus, the overlap metrics in the dataset are comparable to those between unrelated texts. 

To evaluate diversity among narratives corresponding to a proverb, we compute average Jaccard similarity between sets of unigrams for the narratives. This score is 0.107, which is comparable to a value of 0.098 for unigram overlap between pairs of narratives from different proverbs. This suggests a high lexical diversity between narratives. 

\begin{table}[!hbt]
\small
    \centering
    \begin{tabular}{l|c|c}
    \toprule
         \textsc{\textbf{LLM}} & \textsc{\textbf{Acc. (\%)}} $\uparrow$ & \textsc{\textbf{MRR}} $\uparrow$ \\
    \midrule
    Random & 0.40 & 0.024 \\
        Word2Vec & 1.52 & 0.047 \\
         BERT & 0.36 & 0.025 \\ 
         ROBERTA & 1.64 & 0.054 \\
         DistilBERT & 1.92 & 0.053 \\
         ALBERT & 0.40 & 0.025 \\
         Sentence-BERT & 13.44 & 0.217 \\
         GPT-2 & 0.92 & 0.033 \\
         BART & 1.14 & 0.041 \\
         T5 & 2.32 & 0.065 \\
    \bottomrule
    \end{tabular}
    \caption{Proverb retrieval performance using word2vec and off-the-shelf LLMs (`base' versions).}
    \label{tab:LLM_performance_without_finetuning}
\end{table}

\noindent \textbf{Using distributional embeddings}:
We explore if we can retrieve the correct proverb corresponding to a narrative only by using similarity in their distributional representations. 
The similarity between a proverb and a narrative is defined as the cosine similarity between the representation of the proverb and the narrative obtained using word2vec embeddings \cite{mikolov2013efficient} or contextual embeddings from LLMs.
Details of implementation are provided in Appendix~\secref{sec:retrieval_model_implementation_details}.


\noindent 
For this retrieval task, we report the accuracy and Mean Reciprocal Rank of the correct proverb in Table \ref{tab:LLM_performance_without_finetuning}. We note that while all models perform better than random (with Sentence-BERT performing the best), the performance is very low when using out-of-the-box representations. In \secref{sec:experiments}, we explore learning-based methods for the same setup.

\subsection{Data characteristics}
\begin{figure}[!h]
    \centering
    \includegraphics[scale=0.32]{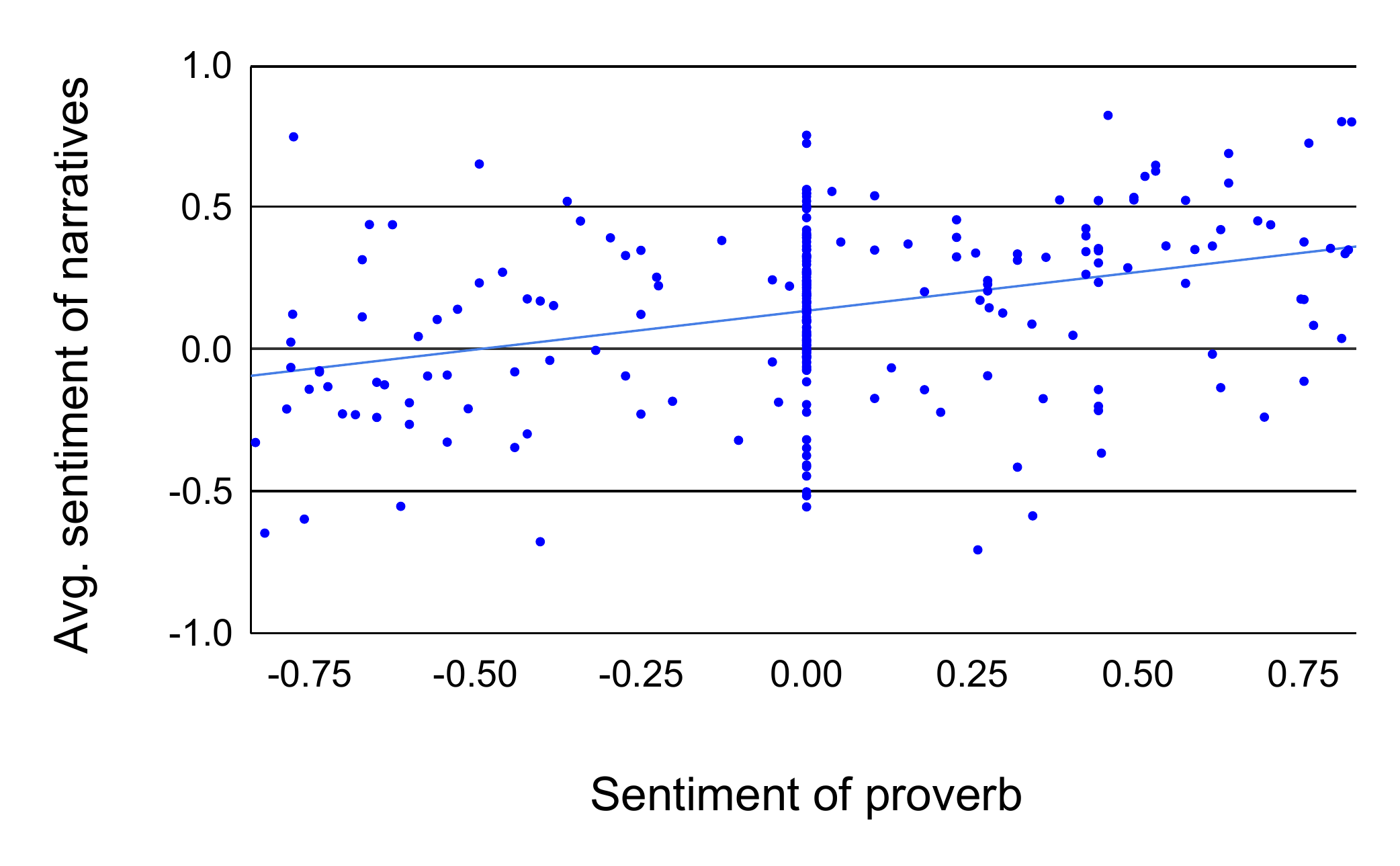}
    \caption{Average VADER sentiment score of narratives corresponding to a proverb against the VADER sentiment score of the proverb. The blue line shows the least-squares fit.} 
    \label{fig:proverbs_sentiment_chart}
\end{figure}

\begin{figure}[!h]
    \centering
    \includegraphics[scale=0.32]{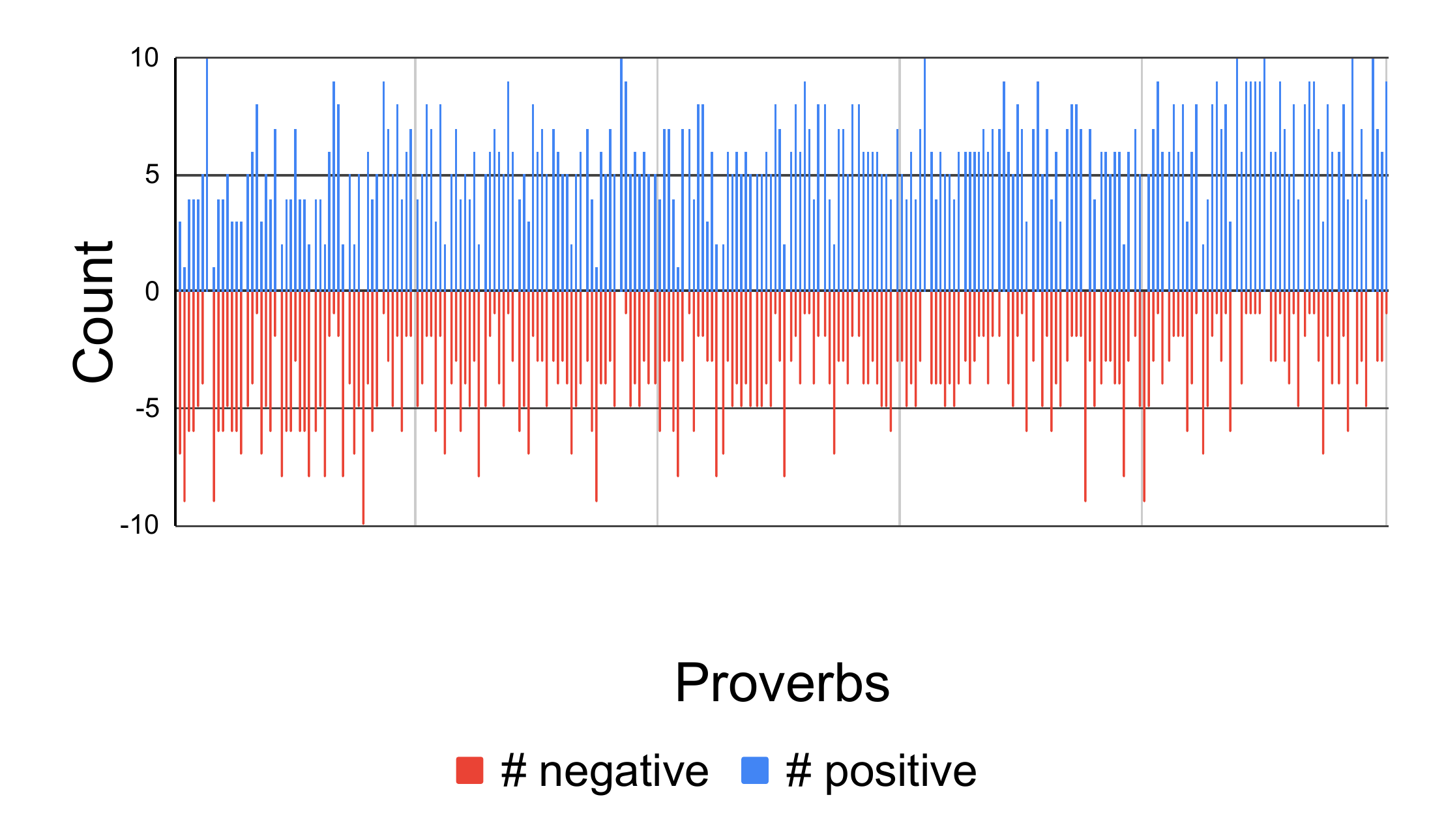}
    \caption{Count of narratives with positive or negative VADER sentiment for each proverb. Proverbs are arranged in increasing order of their own VADER sentiment scores. Neutral sentiment narratives are excluded. For count of negative sentiment narratives (shown in red), consider the absolute value.} 
    \label{fig:sentiment_variance_chart}
\end{figure}

\textbf{Diversity of narrative events:}
\label{sec:diversity-sentiment}
\begin{figure}[!ht]
    \centering
    \includegraphics[width=0.80\columnwidth]{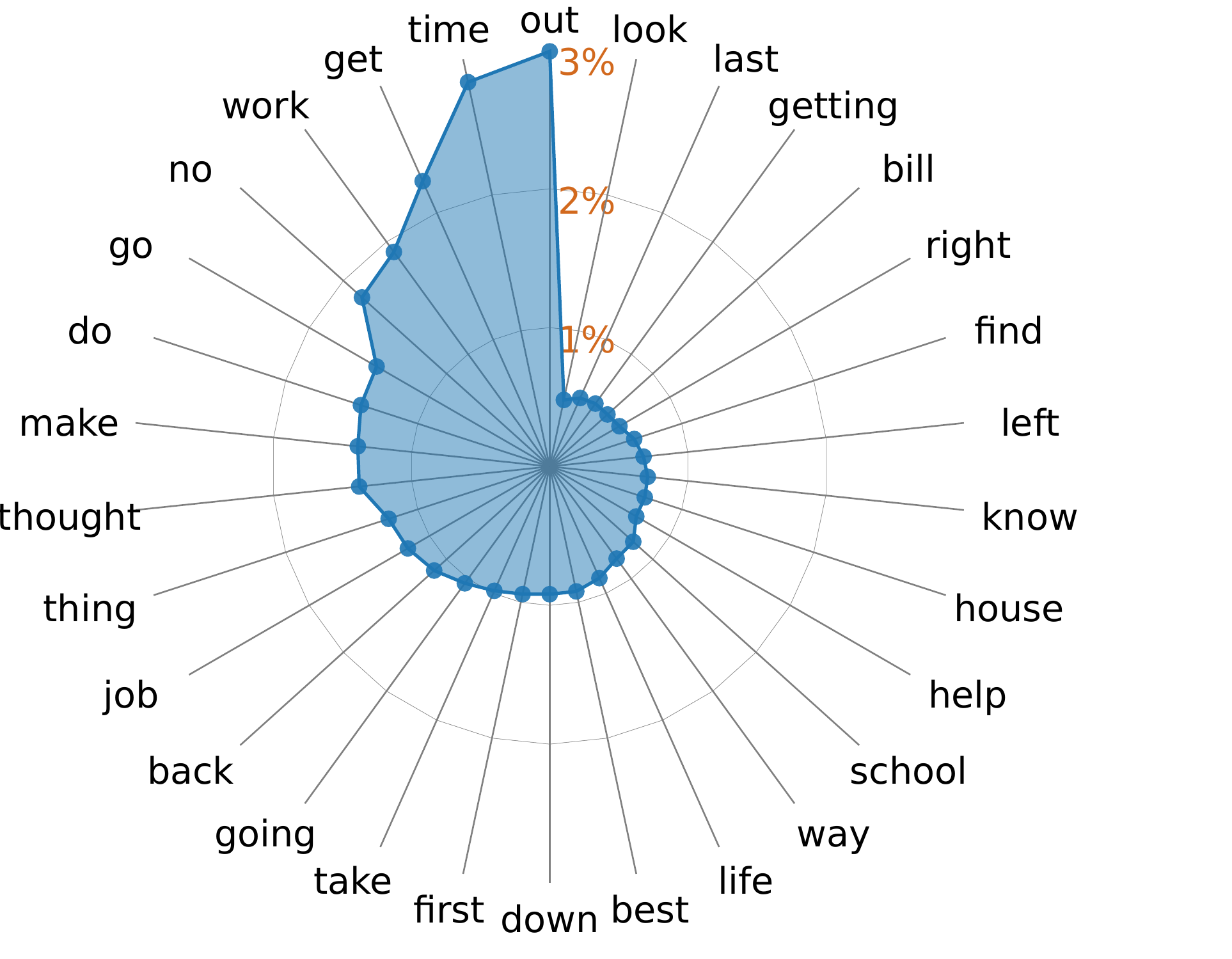}
    \caption{Top-30 `event'/`process' hyponyms in \model.}
    \label{fig:event_hypernym_chart}
\end{figure}
Fig~\ref{fig:event_hypernym_chart} shows the distribution of events in our dataset. Following \citet{roc} we find events as the hyponyms of the word `event' or `process' using WordNet \cite{wordnet}. We see that the top events comprise less than 3\% of all events in our dataset, and the long tail of less frequent events shows the diversity of the dataset. 
\\
\textbf{Sentiment analysis:} 
To evaluate the presence of sentiment association bias between proverbs and corresponding narratives (e.g., if negative sentiment proverbs only correspond to negative sentiments in narratives), we perform sentiment analysis of the narratives using VADER~\cite{hutto2014vader}. Figure~\ref{fig:proverbs_sentiment_chart} shows the average sentiment scores of the narratives corresponding to a proverb plotted against the sentiment score of the proverb. 
We find that the narratives are diverse in terms of their sentiment polarities showing a weak positive correlation (Pearson correlation score 0.35) with the sentiment score of the proverbs. 
Figure~\ref{fig:sentiment_variance_chart} shows the variance in terms of the number of positive and negative sentiment narratives (out of 10) for each proverb, showing a diverse spread of narrative sentiment polarities across proverbs. For additional details, please refer to Appendix~\secref{sec:additional_dataset_analysis}.
\\
\noindent
We perform a few additional analyses on our dataset 
and found that (1) around 61\% of mentions in the narratives were male, (2) diverse spread of reading complexity values in narratives measured using \textit{Fleisch reading ease}\footnote{\url{https://en.wikipedia.org/wiki/Flesch_Kincaid_readability_tests}}, and (3) absence of any hate speech in the narratives of our dataset. The detailed experiments for these analyses are given in Appendix~\secref{sec:additional_dataset_analysis}. 

\begin{table}[!tb]
\small
    \centering
    \begin{tabular}{l|c|c|c}
    \toprule
         \Thead{Criterion} & \Thead{\model} & \Thead{[1]} & \Thead{[2]} \\
         \midrule
         Relatedness & 3.91 & 3.15 & \textbf{3.92}\\
         Interesting/Creative & 3.57 & 3.34 & \textbf{3.63}\\
         Fluency & \textbf{3.98} & 3.23 & 3.80\\
         Overall & \textbf{3.68} & 3.15 & 3.66\\
         \bottomrule
    \end{tabular}
    \caption{Averaged Likert scale ratings for data quality. Overall ratings for \model are better than [1] \citet{quote2020emnlp} and [2] \citet{ltq1}.} 
    \label{tab:quality_check}
\end{table}
\subsection{Human Evaluation of Dataset Quality}
\label{sec:quality}
We perform a human evaluation of the narratives in our dataset on various criteria to judge the quality of our dataset. We perform this evaluation using the AMT platform. We randomly sample 250 proverb-narrative pairs and ask the turkers to evaluate the narratives on the following criteria:
\begin{itemize}[noitemsep, topsep=0pt, leftmargin=*]
\item \textbf{Relatedness:}  how closely the narrative reflects the meaning of the proverb (1: totally unrelated, 5: perfectly related)
\item \textbf{Interesting/Creative:} how much is the narrative like a short creative or interesting story (1: very uninteresting/boring, 5: very creative/story-like)
\item \textbf{Fluency:} grammatical correctness of the narrative
(1: poor English with grammatical mistakes, 5: perfect English with no errors in writing)
\item \textbf{Overall rating}
\end{itemize}
All the ratings are done on Likert scales from 1 to 5, where 1 is the lowest value for each criterion and 5 is the highest. Also, the rating value `3' was calibrated to be slightly leaning to the higher end of the scale (instead of neutral) so that the turkers take a clear stand on the polarity of each criterion. Table~\ref{tab:quality_check} shows the qualitative evaluation of our dataset. The average overall rating was 3.67 and the average pair-wise inter-annotator agreement for labeling a narrative as overall good vs overall poor (overall score >= 3 vs < 3) is 0.84\footnotemark. We also rate the quality of the aligned spans in our dataset similarly on a scale of 1 to 5. The average rating of the alignment between spans was 3.91 and the average pair-wise inter-annotator agreement for alignment as good vs poor (rating >= 3 vs < 3) is 0.86\footnotemark[\value{footnote}].
\footnotetext{\label{kappa_footnote}Due to label imbalance 
kappa statistics for inter-annotator agreement are not reliable \cite{feinstein1990high}. Thus, we report average pairwise agreement score, i.e. how often two judges agree on a label for a sample.
}
\\
\begin{table*}[!ht]
\small
    \centering
    \begin{tabular}{l|c|c|c|c}
    \toprule
         \Thead{Characteristics} & \citet{ltq1} & \citet{ltq3} & \citet{quote2020emnlp} & \model \ \\
         \midrule
         Domain & Fiction & Social Media & Social Media & Fiction \\
         Manual curation of narratives & \xmark & \xmark & \xmark & \cmark \\
         Alignment annotation & \xmark & \xmark & \xmark & \cmark \\
         Focus on proverbs & \xmark & -\tablefootnote[6]{We did not have access to the dataset to verify this.} & \xmark & \cmark \\
         \bottomrule
         
    \end{tabular}
    \caption{Comparing \model with prior works on learning to quote based on different characteristics of the data and the collection process. While previous methods collect contexts and labels by mining existing text resources through heuristics (with no manual curation), \model contains contexts in form of narratives authored by crowdworkers explicitly for this task. \model further provides fine-grained alignment annotation between narratives and proverbs.
    }
    \label{tab:key_differences}
    \vspace{-0.5cm}
\end{table*}
\noindent Table~\ref{tab:key_differences} highlights the key differences between \model and prior work that dealt with related figurative language tasks involving quotes. Notably, \model exclusively deals with proverbs unlike prior work (which includes common phrases and idioms such as ``trust your gut") and also provides granular annotations in form of annotated spans. Also, \model contains narratives crowdsourced by specifically keeping proverbs in focus, rather than obtaining them using heuristic supervision. To quantify dataset quality, we ran human evaluation similar to \model over  (1) 200 randomly drawn samples from the ``Reddit" dataset of quotations in context from the \citet{quote2020emnlp}, and (2) 200 randomly drawn samples from the corpus of \citet{ltq1}. Based on average Likert scores in Table~\ref{tab:quality_check} we find that \model is (1) significantly superior (using t-test; $p < 0.05$) on all criteria than \citet{quote2020emnlp}, and (2) better in overall ratings than \citet{ltq1}.

\section{Tasks \& Evaluation}
\label{sec:experiments}
In this section, we introduce three tasks associated with \model and describe their experimental setup and benchmark results: (1) Proverb and Alignment Prediction, (2) Narrative Generation, and (3) Identifying narratives with similar motifs. 
\vspace{-0.1cm}
\subsection{Proverb and alignment prediction}
\label{sec:proverb_retrieval}
\subsubsection{Task details}
In this task, the objective is to predict the correct proverb for a given narrative from the set of 250 proverbs in the dataset. The motivation of this task is to test whether language models can abstract the underlying meaning of the narratives and make an analogy with the correct proverb from a large set of proverbs. In terms of applications, this task is related to proverb recommendation, which can be useful in creative writing assistants. The task is challenging as there might be multiple proverbs loosely related to the narrative context, but not be completely consonant with subliminal themes in the narrative. An underlying assumption here is that a narrative would match well with exactly one proverb. We found this reasonable for most examples in the dataset. 

\subsubsection{Experiment Setup and Results}
\label{sec:proverb_prediction_setup}
We consider two settings, predicting (1) Seen and (2) Unseen proverbs.
\begin{itemize}[noitemsep, topsep=0pt, leftmargin=*]
    \item Seen proverbs: The set of proverbs in the train and test set are the same. We divide narratives corresponding to each proverb into train and test in 6:4 ratio. So, the train and test sets have 1500 and 1000 proverb-narrative pairs respectively.
    \item Unseen proverbs: Here, we consider 150 proverbs in the train set and the remaining 100 proverbs in the test set (6:4 split on the set of proverbs). The sets of proverbs in the train and test split are disjoint. So, the train and test sets have 1500 and 1000 proverb-narrative pairs respectively (since each proverb has 10 narratives).
\end{itemize}
\textbf{Proverb prediction:}
Here we focus on only predicting the corresponding proverb for a narrative, without employing the span alignments in training or evaluation. For this, 
we fine-tune the retrieval models based on different LLMs previously described in \secref{sec:data_analysis} (details of models in Appendix~\secref{sec:proverb_prediction_model_implementation_details}). 
To evaluate performance we consider accuracy and Mean Reciprocal Rank as metrics. Table~\ref{tab:LLM_performance_with_finetuning} shows best proverb prediction performance on test split for `seen' and `unseen' proverbs\footnote[7]{Our reported accuracies denote the highest accuracy achieved on the test set during model training as we do not have a validation set to choose the best model.}. RoBERTa performs the best for both the `seen' and `unseen' settings, and the performance for all models is consistently lower for unseen proverbs (as would be expected, since this task involves much greater generalization). Further, while the performance of all models is much better than chance, even the highest performance is only 28.2\%.
\begin{table}[!ht]
\small
    \centering
    \begin{tabular}{l|r|r}
    \toprule
        \textsc{\textbf{Model}} & \textsc{\textbf{Acc. (\%)}} $\uparrow$& \textsc{\textbf{MRR}} $\uparrow$ \\
    \midrule
    \rowcolor{LightCyan}
         \multicolumn{3}{c}{\textbf{Seen proverbs}}\\
    Random & 0.4 & 0.024 \\
         BERT & 22.9 & 0.342 \\ 
         RoBERTa & 28.2 & 0.391 \\
         DistilBERT & 18.7 & 0.289 \\
         ALBERT & 13.4 & 0.221 \\
         Sentence-BERT & 20.6 & 0.315 \\
         BART & 15.8 & 0.245  \\
         T5 & 18.7 & 0.292 \\
    \midrule
    \rowcolor{LightCyan}
         \multicolumn{3}{c}{\textbf{Unseen proverbs}}\\
         Random & 1.0 & 0.005 \\
         BERT & 19.2 & 0.307 \\ 
         RoBERTa & 20.3 & 0.314 \\
         DistilBERT & 17.4 & 0.277 \\
         ALBERT & 1.1 & 0.053 \\
         Sentence-BERT & 17.0 & 0.278 \\
         BART & 8.5 & 0.189\\
         T5 & 13.7 & 0.242 \\
    \bottomrule
    \end{tabular}
    \caption{Proverb prediction performance on `seen' and `unseen' proverbs (all LLMs are in `base' version).}
    \label{tab:LLM_performance_with_finetuning}
\end{table}
\\
\textbf{Alignment prediction:}
Here we focus only on predicting an aligned span from the narrative given the narrative, proverb, and a span from the proverb as inputs. We fine-tune two large language models (BERT and RoBERTa) for this by adopting a learning framework similar to answer span prediction for SQUAD~\cite{rajpurkar2016squad}. The language model outputs two probability distributions corresponding to the start and end positions of a span, over the narrative tokens. We iterate over all the combinations of the start and end tokens and choose the span with maximum likelihood. For span prediction, we report token-level precision, recall, and F1. 
\begin{table}[!h]
\small
    \centering
    \begin{tabular}{l|c|c|c}
    \toprule
         \Thead{Model} & \Thead{Span P} & \Thead{Span R} & \Thead{Span F1} \\
         \midrule
         BERT & 0.070 & 0.123 & 0.089\\
         RoBERTa & 0.068 & 0.143 & 0.092\\
         \bottomrule
    \end{tabular}
    \caption{Alignment prediction performance for seen proverbs using LLMs (`base' versions).}
    \label{tab:span_prediction_seen}
    \vspace{-6px}
\end{table}
Table~\ref{tab:span_prediction_seen} shows the results of alignment prediction on the `seen' proverbs using BERT and RoBERTa models. We find that the performance is low for both models indicating major scope for improvements.
\\
\textbf{Predicting proverbs and alignment jointly:} We formulate this as multi-task learning. We extend the models from the proverb prediction task by adding a component to predict span from narrative given a span from the proverb and the narrative. The language model is thus shared across the proverb prediction and span prediction tasks.
The span prediction branch predicts the start and end position of the corresponding narrative span. We jointly train the model with multi-task learning of the two tasks, i.e., proverb and alignment prediction, on the `seen' proverbs data split. We report the 
accuracy for proverb prediction and precision, recall, and F1 for span prediction.
Apart from this joint model, we also consider a pipelined baseline model which first does proverb prediction, followed by span prediction if the correct proverb was predicted.
Table~\ref{tab:joint_seen} shows results for the joint model and the pipelined-baseline model.
The low performance of the models indicates major scope for improvements in the individual tasks.
While in principle the two tasks should benefit from joint training, we find that joint training performs worse than pipelined-baseline for both proverb and alignment prediction. Future work can explore designing better models for joint training to leverage the interdependence between proverb prediction and alignment prediction.

\addtolength{\tabcolsep}{-2pt} 
\begin{table}[!h]
\small
    \centering
    \begin{tabular}{l|c|c|c|c}
    \toprule
         \Thead{Model} & \Thead{Acc. (\%)} & \Thead{Span P} & \Thead{Span R} & \Thead{Span F1} \\
         \midrule
         \rowcolor{LightCyan}
         \multicolumn{5}{c}{\textbf{Pipelined}}\\
         BERT & 22.9 & 0.018 & 0.035 & 0.024\\
         RoBERTa & 28.2 & 0.019 & 0.048 & 0.027\\
         \midrule
         \rowcolor{LightCyan}
         \multicolumn{5}{c}{\textbf{Joint training}}\\
         BERT & 19.8 & 0.015 & 0.029 & 0.019\\
         RoBERTa & 26.5 & 0.015 & 0.030 & 0.020\\
         \bottomrule
    \end{tabular}
    \caption{Joint proverb and alignment prediction performance for seen proverbs using LLMs (`base' versions).}
    \label{tab:joint_seen}
\end{table}
\addtolength{\tabcolsep}{2pt} 
\subsubsection{Qualitative analysis of proverb prediction models}
\label{sec:mcq-human-analysis}
\begin{figure}[!ht]
    \centering
    \vspace{-0.4cm}
    \includegraphics[scale=0.45]{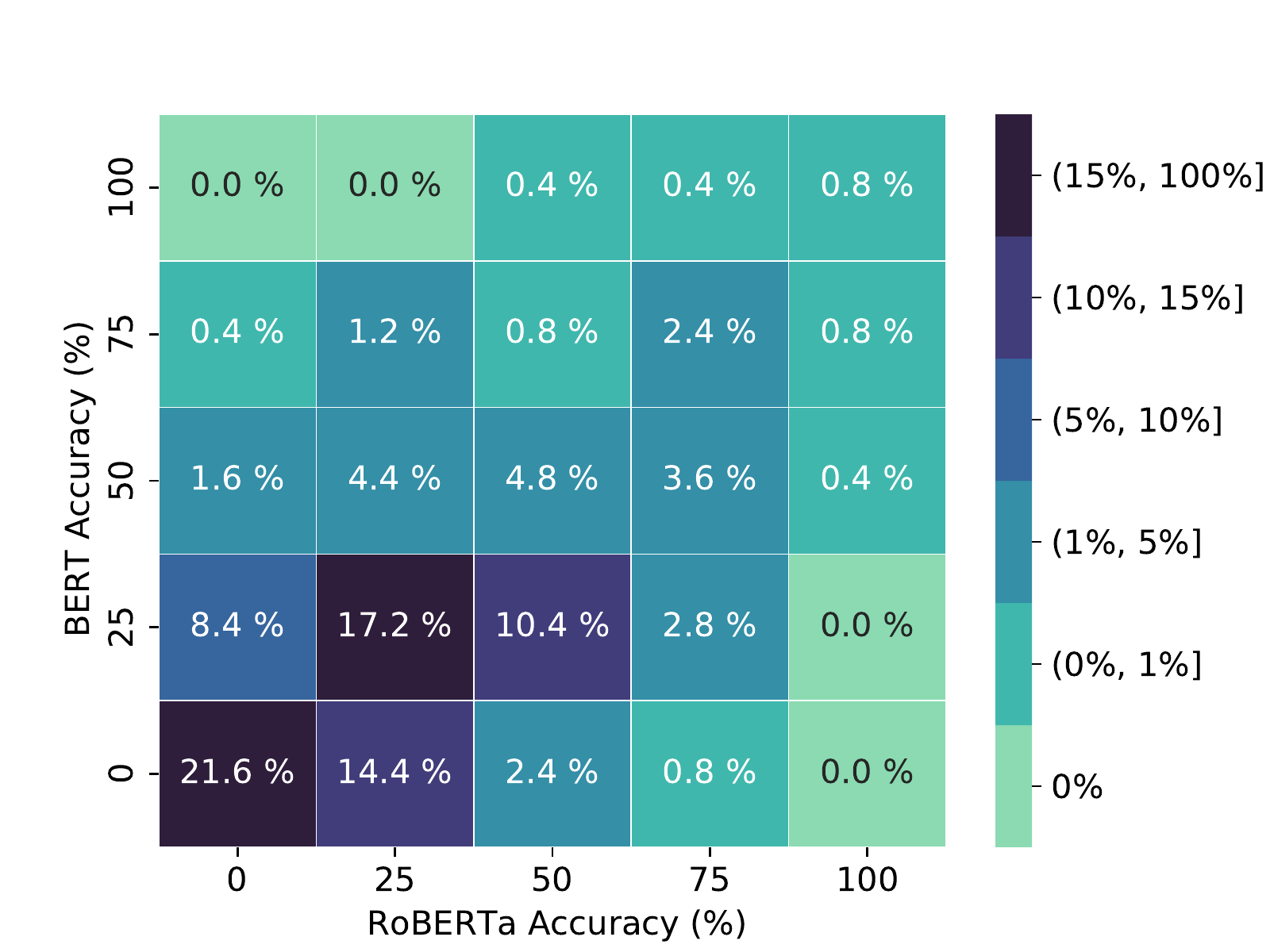}
    \caption{Heatmap showing the percentage of proverbs with various fine-tuned BERT and RoBERTa proverb prediction accuracies (for example,  more than 15\% of the proverbs have RoBERTA prediction accuracy as 25\% and BERT prediction accuracy as 25\%).
    }
    \label{fig:per_proverb_acc}
    \vspace{-0.4cm}
\end{figure}

Figure~\ref{fig:per_proverb_acc} shows a heatmap to study the differences in prediction accuracies of BERT and RoBERTa models.
We see that RoBERTa generally outperforms BERT for many cases (in Figure~\ref{fig:per_proverb_acc},  values in the bottom-right triangle are typically greater than the top-left). 
Looking into the narratives for proverbs in the test set with high accuracy (>=0.75), we think a reason for the high performance could be the presence of certain words/phrases which are synonymous to some words/phrases in the proverb (for example, presence of word `group' for the proverb `\textit{birds of a feather flock together}'). 
\noindent
On the other hand, there are cases when the model is confused because of multiple topics being discussed in the narrative resulting in an incorrect prediction. For example, some narratives in the test set for the proverb `\textit{life's not all beer and skittles}' describe earning money the hard way, which confused the RoBERTa model into predicting `\textit{time is money}' for such narratives.

\subsubsection{MCQ task for human performance comparison}
To formulate a feasible task for humans,  we frame proverb prediction as a multiple choice question (MCQ) task where for each narrative, 5 proverbs are provided as choices. The set of choices includes the correct proverb and 4 other distractor proverbs, chosen by using the fine-tuned RoBERTa model. Examples of the MCQ task and details of choosing distractors are provided in Appendix \secref{sec:human_eval_mcq_appendix}. 
Table~\ref{tab:mcq} shows the accuracy of the human evaluation for this MCQ task on a random sample of 100 narratives from the test split of "seen" proverbs conducted using AMT.
Compared to RoBERTa, we find humans are much better at this adversarially created MCQ task.
Note that the performance for RoBERTa in Table~\ref{tab:mcq} and Table~\ref{tab:LLM_performance_with_finetuning} is different, as Table~\ref{tab:mcq} reports accuracy only on the random sample of narratives chosen for human evaluation. 
The estimate for human performance is likely an under-estimate since in many cases human subjects were unfamiliar with the meanings of some of the proverbs provided in the options and as a result, focused more on surface-level cues (details of this analysis are provided in Appendix~\secref{sec:human_eval_mcq_appendix}).
The average pair-wise inter-annotator agreement between human subjects for this task was 0.73\textsuperscript{\ref{kappa_footnote}}.

\noindent This evaluation does not take into account semantic similarity between proverbs (two proverbs might be equally suitable for the same context). To explore this, we analyze the human errors on the MCQ task and find that in only around 11\% of the errors, the proverb chosen by humans is semantically similar to the annotated proverb and can also be a suitable answer to the MCQ task. Details about this analysis are given in Appendix~\secref{sec:error_analysis_semantically_similar_proverbs}. Future work can consider handling semantic similarity between proverbs explicitly and devise suitable evaluation metrics.

\begin{table}[]
\small
    \centering
    \begin{tabular}{l|c}
    \toprule
         Predictor & \Thead{Acc.(\%)} $\uparrow$ \\
         \midrule
         RoBERTA & 23.0 \\
         Human & 78.7 \\
         \bottomrule
    \end{tabular}
    \caption{Proverb prediction accuracy in MCQ setting.}
    \label{tab:mcq}
    \vspace{-0.5cm}
\end{table}

\subsection{Narrative Generation}
\label{sec:narrative_generation}
\subsubsection{Task details}
One of the important use-cases for NLP models in the creative writing domain is to use these models to generate content. We explore the task of generating narratives corresponding to a proverb and a given topic (specified as a set of keywords). 
We benchmark the performance of two recently proposed state-of-the-art models in text generation, T5~\cite{t5} and BART~\cite{bart}, by fine-tuning them on \model. 

\subsubsection{Experiments and Results}
We divide our dataset into train and test split under `seen' and `unseen' proverbs settings similar to the proverb prediction task. 
We consider the set of verbs and named-entities as the keywords for a narrative. We train our narrative generation model conditioned on the proverb and the keywords.

\noindent Table~\ref{tab:narrative_generation_automatic_seen} shows results for automatic evaluation of the generated narratives using BLEU~\cite{papineni2002bleu}, ROUGE-L~\cite{lin2004rouge}, and recall of the keywords mentioned in the generated narrative as metrics. Examples of generated narratives are given in Appendix \secref{sec:generated_narratives_appendix}. We find that BART performs better than T5 on the automatic evaluation metrics.
Further, we perform human evaluation to evaluate the quality of the generated narratives in AMT by considering the same criteria (and rating semantics) employed in Section 4.3. 
Table~\ref{tab:narrative_generation_result} shows the human evaluation of generated narratives using BART and T5 when tested over `seen' proverbs. Low scores for BLEU and ROUGE-L in automatic metrics and low Likert ratings of the generated narratives indicate much scope for future improvement on this task.

\begin{table}[!ht]
\small
    \centering
    \begin{tabular}{l|c|c|c}
    \toprule
         \Thead{Model} & \Thead{BLEU} $\uparrow$& \Thead{ROUGE-L} $\uparrow$& \Thead{Recall}  $\uparrow$\\
         \midrule
         \rowcolor{LightCyan}
         \multicolumn{4}{c}{\textbf{Seen proverbs}}\\
         BART & 4.21 & 30.80 & 0.90\\
         T5 & 2.25 & 27.83 & 0.77 \\
         \midrule
         \rowcolor{LightCyan}
         \multicolumn{4}{c}{\textbf{Unseen proverbs}}\\
         BART & 4.39 & 31.36 & 0.93 \\
         T5 & 2.34 & 26.61 & 0.75 \\
         \bottomrule
    \end{tabular}
    \caption{Automatic evaluation for narrative generation on `seen' and `unseen' proverbs using `base' versions of LLMs.}
    \label{tab:narrative_generation_automatic_seen}
    
\end{table}
\begin{table}[!ht]
\small
    \centering
    \begin{tabular}{lcc}
        \toprule
        \Thead{Criterion} & BART & T5  \\
         \midrule
         Relatedness & 2.75 & 2.57\\
         Interesting/Creative & 2.97 & 3.07 \\
         Fluency & 2.71 & 2.53 \\
         Overall & 2.87 & 2.76 \\
         \bottomrule
    \end{tabular}
    \caption{Human evaluation results for narrative generation on `seen' proverbs.}
    \label{tab:narrative_generation_result}
\end{table}

\subsection{Identifying narratives with similar motifs}
\label{sec:discovering_similar_narratives}
\subsubsection{Task details}
An important aspect of language understanding is the ability to make linguistic (and narrative) analogies, i.e., identifying `similarity' between narratives (e.g., identifying two narratives that are variations on the `Cinderella story' theme).
Here, we explore the task of identifying narrative analogy by modeling `similarity' between narratives based on proverbs illustrated by them.
For this task, two narratives are taken to be similar if they are related to the same proverb. 

\subsubsection{Experiments and Results}
For this task, we use the train and test split of `seen' proverbs setup in the proverb prediction task. The aim is to find similar narratives for each narrative in the test split amongst all narratives in the test split. So for each narrative, there are 3 other similar narratives (corresponding to the same proverb) in the test split (containing 1000 narratives).

\paragraph{Modeling similarity between narratives} We use the learned models in the proverb prediction task to obtain a probability distribution over the proverbs for each narrative. To model similarity, we compute the distance between the (vectors representing) two probability distributions using one of the following: (1) cosine distance; (2) Jenson-Shannon divergence; (3) L2 (Euclidean) distance; and (4) L1 (Manhattan) distance. We predict the narrative closest (in terms of distance metrics) to the input narrative as the most similar.
Table~\ref{tab:discovering_similar_narratives} shows the accuracy of getting a similar narrative using different distance metrics and different fine-tuned LLMs. Using cosine or Jenson-Shannon divergence as the distance metric on the probability distribution over proverbs predicted by the RoBERTa model performs best on this task. However, the overall performance of models are still low and can be benefited by devising suitable training methods for this task.

We perform an additional experiment on finding similar narratives without performing proverb prediction as an intermediate step. We use a pre-trained Sentence-BERT model to obtain representations of each narrative. For a given input narrative, we calculate the cosine distance between the Sentence-BERT representation of the input narrative and all other narratives in the test set. We predict the narrative having minimum cosine distance to the input narrative as the most similar. Using this approach we find the accuracy of identifying similar narratives as 6.6\%, which is lower than most values reported in Table~\ref{tab:discovering_similar_narratives}. This low value highlights the diversity between narratives and the challenge in finding analogies between narratives.

\begin{table}[!h]
\small
    \centering
    \begin{tabular}{l|r|r|r|r}
    \toprule
         \Thead{Model} & \Thead{Cos} & \Thead{JSD} & \Thead{L2} & \Thead{L1}  \\
         \midrule
         BERT & 8.5 & 8.0 & 7.3 & 7.9 \\
         RoBERTa & 13.3 & 13.4 & 11.2 & 11.8\\
         Distil-BERT & 6.5 & 7.2 & 5.2 & 6.0\\
         Sentence-BERT & 7.2 & 6.1 & 7.0 & 5.9\\
         \bottomrule
    \end{tabular}
    \caption{Prediction accuracy (\%) for identifying similar narratives 
    by using different distance metrics and distribution over proverbs from different LLMs (`base' versions).}
    \label{tab:discovering_similar_narratives}
\end{table}
\vspace{-0.1cm}
\section{Conclusion and Future Work}
\label{sec:conclusion}
We introduce \model, a high-quality crowdsourced dataset of narratives paired with proverbs, and a suite of challenging tasks associated with this dataset. We show that these provide a challenging testbed for evaluating abstract reasoning and analogical abilities of LLMs. 
Future work can explore more sophisticated mechanisms to use alignment annotations in improving the performance for proverb prediction and model interpretability. Additionally, researchers can explore conditional narrative generation 
through more informative prompts than using keywords. 
\model can also be extended in the future by incorporating more proverbs and adding more layers of complexity like sarcasm or adversarially creating harder narratives. Most of all,
the development of similarly challenging resources and tasks can enable the possibility of socially grounded NLP systems. 

\section*{Ethics and Broader Impact}
\label{sec:ethics}

In \secref{sec:data_analysis}, we note that our dataset shows considerable differences in the distribution of gender of entities (61\% male vs 39\% female), whereas in the real world we expect the ratios to be about equally balanced. Systems that don't account for this bias might end up performing better for narratives with male entities than with females. However, we note that narratives with male and female entities show no differences in overall length or the average number of mentions to those entities.

The proverbs used in our dataset were collected from free public resources without violating intellectual property rights. We do not collect any personal information from the turkers who participated in our crowdsourced tasks. We release our dataset publicly without mentioning any personal details of turkers available automatically in AMT (such as turker IDs). The turkers were compensated fairly and the payment per task is equivalent to an hourly compensation that is greater than minimum wage (based on the median time taken by turkers).

For all the crowdsourcing tasks in this work, we limited the locale of eligible turkers to the USA, Canada, and the UK. Further, to ensure good-faith turkers, we required that the approval rate of the turkers be above 97\%. 

Our screening process has selection biases that likely over-samples narrative-writers from demographics that are over-represented on AMT (ethnically white, college-educated, lower-to-medium income, and young)~\cite{hitlin20164}, and this is likely to have affected the topics and type of language usage in the collected narratives.

Finally, our investigation here has focused on traditional English proverbs, even while proverbs are universal in human languages and cultures~\cite{penfield1988proverbs}. This poses a real risk of the development of AI models that better understand and employ specific types of figurative language than others. Such systems are likely to be less user-friendly to users that don't belong to specific social-cultural backgrounds. To mitigate these risks, but also since proverbs are universal repositories of culture-specific knowledge, future work should extend our effort to more equitably represent the variety and diversity of human thought and cultural experiences. Our investigation here, unfortunately, does not adequately do this. As the proverb goes, the road to hell is paved with good intentions.


\bibliography{anthology,custom}
\bibliographystyle{acl_natbib}
\appendix
\section*{Appendix}
\section{Additional dataset analysis}
\label{sec:additional_dataset_analysis}
\noindent \textbf{Additional details on sentiment analysis:}
An example of proverb for which the narratives were close in sentiment scores to the proverb is `\textit{a thing of beauty is a joy forever}' while for `\textit{there's no fool like an old fool}' the sentiment polarity of narratives was on average opposite to that of the proverb. 
We note that there are indeed a small number of proverbs for which all or most narratives leaning towards a particular sentiment polarity. Quantitatively, for 23 proverbs, either 9 or all 10 of the narratives have positive VADER sentiment score. These include: `\textit{Nothing succeeds like success}' , `\textit{Christmas comes but once a year'} and `\textit{Genius is one percent inspiration, ninety-nine percent perspiration}'. There are 6 proverbs for which either 9 or all 10 narratives have a negative VADER sentiment score. These include: `\textit{The wages of sin is death}', `\textit{Fish always stink from the head down}' and `\textit{Don't wash your dirty linen in public}'. 
However, as seen in Figure~\ref{fig:sentiment_variance_chart}, the vast majority of proverbs in the dataset are represented by narratives with both positive and negative sentiment polarities. 

\noindent \textbf{Gender distribution of entities:} 
Using an off-the-shelf neural coreference pipeline, we find that 61\% of the mentions in the narratives are male, while 39\% are female. Around 48\% of the narratives have predominantly male mentions, 26\% of the narratives have predominantly female mentions and the rest have equal number of male and female mentions. The average number of words in predominantly male and female mention containing narratives was comparable (~65 words).
\\
\textbf{Language complexity:}
We use the \textit{Fleisch reading ease}\footnote{\url{https://en.wikipedia.org/wiki/Flesch_Kincaid_readability_tests}} to calculate language complexity of narratives in our dataset.
The reading scores vary from 112.1 (equivalent to 3rd grade reading levels) to -41.5 (significantly above college graduate reading levels) with an average score for the narratives in our dataset as 66.5 (equivalent to 8th/9th grade reading levels), showing a considerable spread in the complexity of language in our dataset.
\\
\textbf{Hate speech:} 
Using an off-the-shelf hate speech classifier 
\cite{davidson2017automated}, we found no instances of hate or toxic speech in the dataset.
 
\section{Human evaluation on MCQ task}
\label{sec:human_eval_mcq_appendix}
We formulated a MCQ task for proverb prediction to gauge human performance. The MCQ task has 5 options -- correct proverb and 4 distractor proverbs. The distractor proverbs were chosen using the fine-tuned RoBERTa model on the proverb prediction task. We choose the distractor proverbs from a mix of proverbs with the highest prediction probabilities, and proverbs that are assigned the most similar probabilities to the correct answer from the RoBERTa model. We performed this study using the Amazon Mechanical Turk platform. We observed that this task is not that simple even for humans and requires a certain level of proficiency in English language or in proverbs specifically. The task is more challenging since the options other than the correct choice in the MCQ task were chosen by picking the most confusing options deemed by the RoBERTa \cite{roberta} model. However, we find that these wrong choices are confusing for humans too. This is because superficially these wrong choices also seem quite related to the narrative and it requires good reasoning skills to identify the correct narrative. The other situation where the turkers failed was when the options contained multiple proverbs which are quite close in meaning. For example, when the options contained both `\textit{there's no accounting for tastes}' and `\textit{Beauty is in the eye of the beholder}' the turkers often chose the former when the annotated proverb was the latter. Table~\ref{tab:mcq_turker_fails} shows examples of narratives along with the choices of proverbs where turkers failed to identify the correct proverb.

\begin{table*}[!ht]
    \centering
    \begin{tabular}{l}
    \toprule
          \textbf{Narrative 1:} \\
         She had been so happy when he had asked her to marry him but three years on, \\
         it seemed that he had so many excuses for not setting a date that she thought that \\
         it was never going to happen. Her happiness eventually 
         turned to despair and \\she considered breaking the engagement.\\
         \\
         \textbf{Choice A} : You win some, you lose some \\
         \textbf{Choice B} : Jam tomorrow and jam yesterday, but never jam today  (\textit{Correct}) \\
         \textbf{Choice C} : Cowards may die many times before their death \\
         \textbf{Choice D} : The course of true love never did run smooth (\textit{Marked})\\
         \textbf{Choice E} : Nothing is certain but death and taxes \\
         \midrule
         \textbf{Narrative 2:} \\
         She didn't want to embarrass her friend when she asked her, "It’s beautiful, \\ isn’t it?" She looked at her friend’s new car and nodded her head in agreement. \\ It was purple, the worst car colour she had ever seen, but she faked a smile \\and congratulated her.\\
         \textbf{Choice A} : Imitation is the sincerest form of flattery \\
         \textbf{Choice B} : From the sublime to the ridiculous is only one step  \\
         \textbf{Choice C} : There's no accounting for tastes (\textit{Marked})  \\
         \textbf{Choice D} : Beauty is in the eye of the beholder  (\textit{Correct}) \\
         \textbf{Choice E} : All publicity is good publicity \\
         
         \bottomrule

    \end{tabular}
    \caption{Tricky MCQ questions from human evaluation task of proverb prediction: The above samples show the challenges in the human evaluation task. In case of narrative 1, the turkers often confuse with choice D which superficially seems related but is not correct. For narrative 2, the proverbs in choices C and D are quite close in meaning, thus resulting in a wrong choice by turkers.}
    \label{tab:mcq_turker_fails}
\end{table*}

\section{Semantically similar proverbs}
\label{sec:error_analysis_semantically_similar_proverbs}
Our chosen set of 250 proverbs in \model includes instances of proverbs that are semantically very similar, or even paraphrases (e.g., `\textit{never judge a book by its cover}' and `\textit{appearances can be deceptive}'). This can be problematic since the presence of semantically similar proverbs as different options in MCQ (and as different classes in proverb classification task) can confuse both humans and automated models. To estimate the extent of this phenomenon, we
perform an analysis of human errors on the aforementioned MCQ task. Out of 64 errors we find that for 20 cases, the chosen proverb was completely unrelated to the actual answer. For 29 out of the remaining 44 cases, the chosen proverb \textit{seems} related to the narrative at first glance, but is not aligned and thus not the best fit. For the remaining 15 cases (23\% of human errors), the chosen proverb would have been equally appropriate for the narrative. Further, in 7 out these 15 cases (11\% of human errors), the chosen proverb is also semantically similar to the annotated proverb.
Future work can consider handling semantic similarity between proverbs explicitly and devise suitable evaluation metrics.

\section {Generated Narratives}
\label{sec:generated_narratives_appendix}
We show some examples of the narratives generated by the BART \cite{bart} and T5 \cite{t5} models for the narrative generation task in Table~\ref{tab:example_narratives}. We see that even though the models try to mention all the keywords but they are not able to generate a coherent narrative.

\begin{table*}[!ht]
    \centering
    \begin{tabular}{l}
    \toprule
          \textbf{Proverb 1:} \\
          Absence makes the heart grow fonder \\
        
          \textbf{Gold narrative:} \\
          Although they never got along together much, she somehow always looked forward \\ to the holidays each year. \\
        
          \textbf{Keywords:} \\
          got, looked, the holidays \\
          
          \textbf{Generated narrative by BART:} \\
          Polly got back into the swing of things, she looked forward to the holidays each year.
          \\
          \textbf{Generated narrative by T5:} \\
          Although he got through the holidays each year, he looked over everything \\
          again to his friends, family and friends.
            \\
          \midrule
          
          \textbf{Proverb 2:} \\
          Those who do not learn from history are doomed to repeat it\\
          
          \textbf{Gold narrative:} \\
          Bob cheated on his girlfriend in college.  He was heartbroken when she broke up with \\ him 
          but moved on.  A few years later he married and 
          cheated on his wife.  She left him, \\ just as his 
          girlfriend did in college.
          \\
          
          \textbf{Keywords:} \\
          broke, did, a few years later, cheated, moved, left, married, heartbroken, Bob \\
          \textbf{Generated narrative by BART:} \\
          Bob cheated on his girlfriend with her best friend.
          He was heartbroken when she broke \\ up with him and moved out of his apartment. 
          A few years later,he got married to the \\ girl he cheated 
          on with, and left his apartment with her.
          He never did again.
          \\
          
          \textbf{Generated narrative by T5:} \\
          Bob was heartbroken. a few years later, he broke 
          into town. He moved out in his life  and \\ moved 
          out.. a few years later, he married a girl.
          A few years later, she cheated on him but,\\
 unfortunately, she left him because she did not do well.\\
    
            \midrule
            
            \textbf{Proverb 3:} \\
            A trouble shared is a trouble halved\\
            
            \textbf{Gold narrative:} \\
            Mike knows that this new problem is going to 
            need some serious help. He tells his friend \\
            Josh about the new problem,and Josh agrees 
            to help him solve it.
            \\
            
            \textbf{Keywords:} \\
            help, solve, knows, going, Mike, Josh, need, tells, agrees
            \\
            
            \textbf{Generated narrative by BART:} \\
            
            Mike knows a thing or two about what is going on in his life that he need help with \\ anything. he tells his friend josh about it and Josh agrees to help solve the problem.
            \\
            \textbf{Generated narrative by T5:} \\
            
            Mike, Josh, knows that he is going to need help to solve the problem. He tells me that \\ he agrees but he will not help me solve the problem.
            \\
          \bottomrule

    \end{tabular}
    \caption{Examples of generated narratives by BART and T5 on `seen' proverbs}
    \label{tab:example_narratives}
\end{table*}


          

          

          



\section{Evaluation of alignment prediction for jointly trained models}
In \secref{sec:proverb_retrieval}, we present models to predict proverb and alignment jointly. During evaluation, we first perform proverb prediction and then perform alignment prediction if the correct proverb was predicted. If an incorrect proverb is predicted, the span precision, span recall, and span F1 are considered as zero.

If we isolate the layers responsible for alignment prediction from the joint model and evaluate it solely on the task of alignment prediction, the span F1 scores for BERT (base) and RoBERTa (base) are 0.078 and 0.074 respectively.

\section{Training details}
In this section we discuss about the model parameters, hyper-parameter settings and hardware and software specifications of training. 

\subsection{Retrieval models' implementation details}
\label{sec:retrieval_model_implementation_details}
As discussed in \secref{sec:lexical}, we formulate a retrieval task to explore if we can retrieve the correct proverb corresponding to a narrative only by using similarity in their distributional representations. We define similarity between a proverb  and a narrative by using cosine similarity between the embeddings of the proverb and the narrative. 
We use (1) word2vec embeddings \cite{mikolov2013efficient} (2) contextual embeddings from LLMs to represent the proverb and narrative. We obtain the embeddings for a context $c$ (where $c$ can be a proverb or a narrative) as:
\begin{itemize}[noitemsep, topsep=0pt, leftmargin=*]
    \item Word2vec: average of word embeddings for tokens in $c$.
    \item BERT~\cite{bert}/RoBERTa~\cite{roberta} : $[CLS]$
    token embedding on passing $c$ through BERT/RoBERTa.
    \item DistilBERT~\cite{distilbert}/AlBERT~\cite{albert} : $[CLS]$ token embedding on passing $c$ through DistilBERT/AlBERT
    \item SentenceBERT~\cite{reimers2019sentence} : normalized SentenceBERT embeddings obtained by using `all-mpnet-base-v2'\footnote{\url{https://huggingface.co/sentence-transformers/all-mpnet-base-v2}} model on $c$.
    \item T5~\cite{t5}/GPT-2~\cite{radford2019language} Encoder: sum of embeddings of tokens in $c$ after passing through the encoder
    
\end{itemize}

\subsection{Proverb prediction models' implementation details}
\label{sec:proverb_prediction_model_implementation_details}
We use the same LLM models (and implementations) used for the retrieval setup discussed in \secref{sec:lexical} and \secref{sec:retrieval_model_implementation_details}.

\subsection{Obtaining keywords for narrative generation}
We consider the named entities and verbs present in a narrative (extracted using \texttt{spacy}~\cite{spacy}) as keywords for generating that narrative.

\subsection{Model parameters}
Our proverb prediction models do not introduce any additional parameters over the existing parameters in the large language models. For joint prediction of proverb and span, we introduce new fully connected layers over the language models, thus introducing ~0.6 M additional parameters.  

\subsection{Hyper-parameter settings}
For all the transformer based models we use the implementation of HuggingFace library \cite{huggingface}. All the model based hyper-parameters are thus kept default to the settings in the HuggingFace library. We use the publicly available checkpoints to initialise the pre-trained models (for example ``bert-base-uncased" checkpoint for initialising BERT\cite{bert}).
For the proverb prediction models we did not truncate any tokens from the proverb and considered the maximum length of the narrative sequence to be 256 tokens. For the alignment prediction and joint training models, we considered the maximum length of the narrative sequence as 230 tokens.
We used the AdamW \cite{adamw} optimizer commonly used to train these models except for T5 \cite{t5}. We used AdaFactor\cite{adafactor} to train our T5 based proverb prediction model. We kept the learning rate as 0.00002 for training. Batch sizes was kept as 16 except for T5, for which we reduced the batch size to 4. The random seed for all experiments was 42. The proverb prediction models were trained for 25 epochs. The BART narrative generation model was trained for 15 epochs and loss converged after that. T5 took longer and was trained for 25 epochs.

\subsection{Software and hardware specifications}
All the models are coded using Pytorch 1.4.0\footnote{\url{https://pytorch.org/}} \cite{pytorch} and related libraries like numpy \cite{numpy}, scipy \cite{scipy} etc. 
We run all experiments on GeForce RTX 2080 GPU of size 12 GB. The system has 256 GB RAM and 40 CPU cores. The proverb prediction models typically take ~2-5 mins for one epoch. For the joint proverb and span prediction models it took roughly 10 mins for one epoch. For narrative generation models it takes 10 mins for BART and around 18 mins for T5 to complete one epoch of training.



\end{document}